\documentclass[letterpaper, 10 pt, conference]{ieeeconf}  % Comment this line out if you need a4paper

\IEEEoverridecommandlockouts                              % This command is only needed if 
\usepackage{times}
\usepackage[pdftex]{graphicx}
\usepackage[ruled,linesnumbered]{algorithm2e}

\makeatletter
\newcommand{\removelatexerror}{\let\@latex@error\@gobble}
\makeatother

\usepackage{subfigure}
\usepackage{amsmath,amssymb,amsopn,amstext,amsfonts}
\usepackage{cancel}
\usepackage[space]{cite}
\usepackage{pdfsync}
\usepackage{balance}
\usepackage{color}
\usepackage{mathtools}
\usepackage{bm}

\usepackage{diagbox}
\usepackage{float}
\usepackage{epstopdf}
\usepackage{pifont}
\usepackage{fixltx2e}
\usepackage{amsmath}
\usepackage{multirow}
\usepackage{url}
\usepackage{verbatim}
\usepackage[linkcolor=black,citecolor=black,urlcolor=black,colorlinks=true]{hyperref}
\usepackage{soul,xcolor}
\usepackage{overpic}
\usepackage{bbding}
\usepackage{amsmath,amssymb}
\usepackage{makecell}

\graphicspath{{../img/}}
\DeclareGraphicsExtensions{.png,.jpg,.eps,.pdf,.jpeg}

\title{\LARGE \bf
FC-Planner: A Skeleton-guided Planning Framework for \\ Fast Aerial Coverage of Complex 3D Scenes
}

\author{Chen Feng$^{2}$, Haojia Li$^{2}$, Mingjie Zhang$^{1}$, Xinyi Chen$^{2}$, Boyu Zhou$^{1,\dag}$, and Shaojie Shen$^{2}$
\thanks{$^{1}$School of Artificial Intelligence, Sun Yat-Sen University, Zhuhai, China.}
\thanks{$^{2}$Department of Electronic and Computer Engineering, The Hong Kong University of Science and Technology, Hong Kong, China.}
\thanks{ {\tt\footnotesize $\{$cfengag, hlied, xchencq, eeshaojie$\}$@ust.hk},}
\thanks{{\tt\footnotesize zagerzhang@gmail.com, zhouby23@mail.sysu.edu.cn}}
\thanks{\textbf{$^{\dag}$ Corresponding Author}}
}

\begin{document}

\maketitle
\thispagestyle{empty}
\pagestyle{empty}

%%%%%%%%%%%%%%%%%%%%%%%%%%%%%%%%%%%%%%%%%%%%%%%%%%%%%%%%%%%%%%%%%%%%%%%%%%%%%%%%
\begin{abstract}

3D coverage path planning for UAVs is a crucial problem in diverse practical applications.
However, existing methods have shown unsatisfactory system simplicity, computation efficiency, and path quality in large and complex scenes.
To address these challenges, we propose \textbf{FC-Planner}, a skeleton-guided planning framework that can achieve fast aerial coverage of complex 3D scenes without pre-processing.
We decompose the scene into several simple subspaces by a skeleton-based space decomposition (SSD).
Additionally, the skeleton guides us to effortlessly determine free space.
We utilize the skeleton to efficiently generate a minimal set of specialized and informative viewpoints for complete coverage.
Based on SSD, a hierarchical planner effectively divides the large planning problem into independent sub-problems, enabling parallel planning for each subspace.
The carefully designed global and local planning strategies are then incorporated to guarantee both high quality and efficiency in path generation.
We conduct extensive benchmark and real-world tests, where \textbf{FC-Planner} computes over 10 times faster compared to state-of-the-art methods with shorter path and more complete coverage. 
The source code will be made publicly available to benefit the community\footnote[3]{\href{https://github.com/HKUST-Aerial-Robotics/FC-Planner}{https://github.com/HKUST-Aerial-Robotics/FC-Planner}}.
Project page: \urlstyle{tt}\url{https://hkust-aerial-robotics.github.io/FC-Planner}.

\end{abstract}

%%%%%%%%%%%%%%%%%%%%%%%%%%%%%%%%%%%%%%%%%%%%%%%%%%%%%%%%%%%%%%%%%%%%%%%%%%%%%%%%
\section{Introduction}
\label{sec:intro}
% background
Recently, Unmanned Aerial Vehicles (UAVs) have become increasingly popular to gather information of target scenes for various tasks, such as structural inspection \cite{bircher2015structural, cao2020hierarchical} and 3D reconstruction \cite{10160933}, due to their agility and flexibility. 
To accomplish these tasks, UAVs need to find a shortest path to fully cover 3D scenes, which is typically formulated as a 3D coverage path planning problem \cite{galceran2013survey}.

% problem
Existing works \cite{bircher2015structural, cao2020hierarchical, jing2016sampling, almadhoun2018coverage, jing2019coverage, tong2022uav,huang2023bim, roberts2017submodular} commonly address this problem by dividing it into two sub-tasks: 1) identifying a set of viewpoints that cover the target scene, 2) determining the shortest path with complete coverage.
However, current solutions for these sub-tasks have limitations that hinder the attainment of satisfactory system simplicity, computation efficiency, and path quality.
For the first sub-task, existing methods require pre-processing using extra tools (\textit{e.g.}, Blender \cite{blender}, CloudCompare \cite{cloudcompare}) to extract free space for ensuring the safety of viewpoints, which increases system complexity. 
Moreover, there is a dearth of an efficient strategy that can produce a minimal set of informative viewpoints comprehensively covering the target scene.
Existing approaches suffer from a trade-off between viewpoints' number and completeness, \textit{i.e.}, a small number of viewpoints cannot guarantee complete coverage while large amount incurs significant computational costs.
On the other hand, the path planner in most frameworks globally optimizes a path fully covering the scene with minimum length, typically formulated as a Combinatorial Optimization Problem (COP) \cite{papadimitriou1998combinatorial}.
However, due to the large scale of the planning problem, existing methods frequently struggle to find high-quality solutions promptly in large and complex scenes. 
Presently, there is a pressing demand for an efficient planning approach in the context of complex 3D scene coverage, which should be capable of decomposing the formidable planning problem into parallelizable sub-problems, ensuring both solution quality and computational efficiency.

\begin{figure}[t]
	\begin{center}
      \includegraphics[width=0.99\columnwidth]{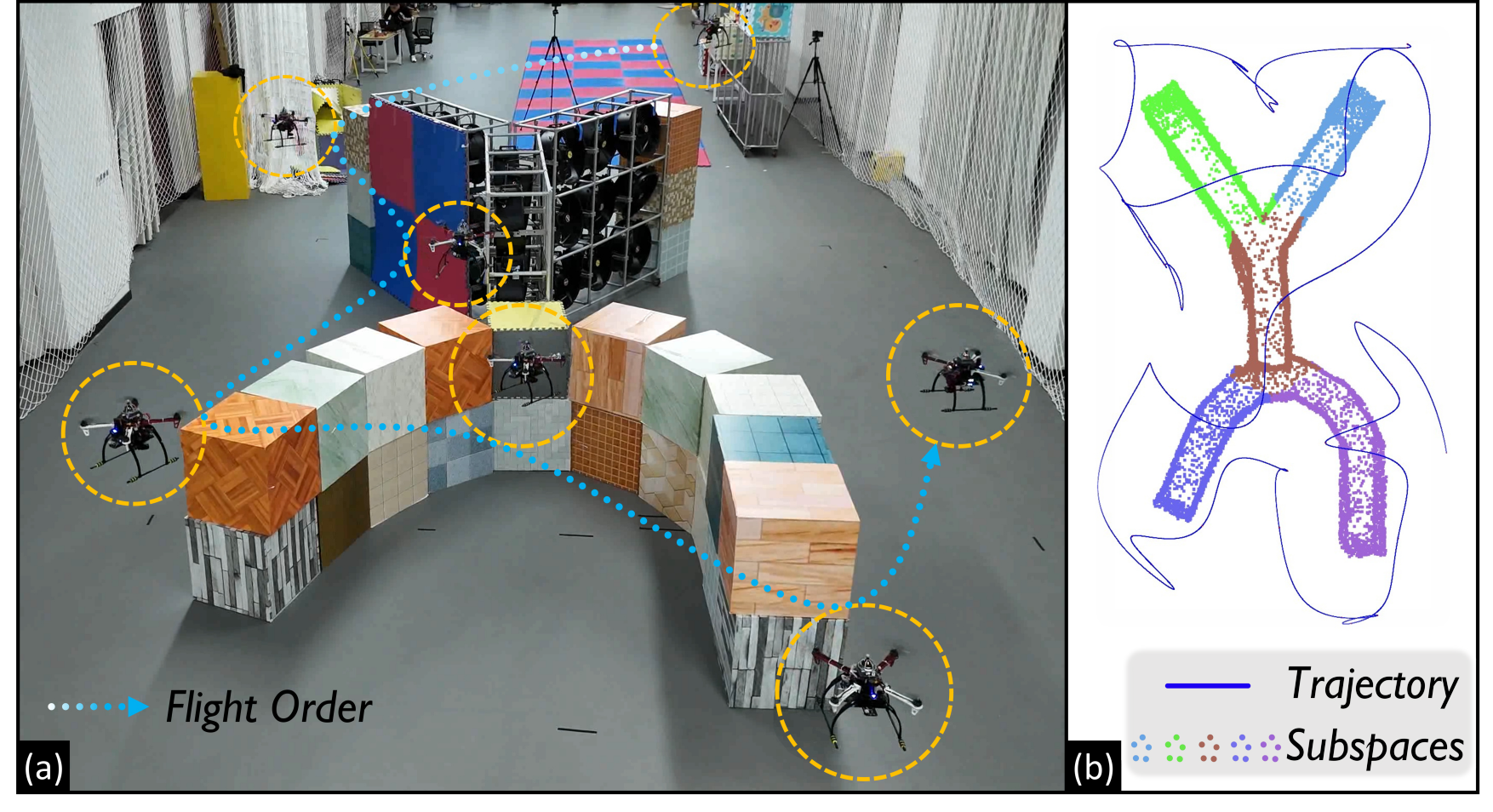}     
      \vspace{-1.0cm}
	\end{center}
   \caption{\label{fig:top} (a) An aerial coverage test conducted in a challenging scene. (b) Showcase of space decomposition details and coverage trajectory.}
   \vspace{-1.5cm}
\end{figure} 

\begin{figure*}[t]
	\begin{center}        
		\vspace{0.1cm}
      \includegraphics[width=1.95\columnwidth]{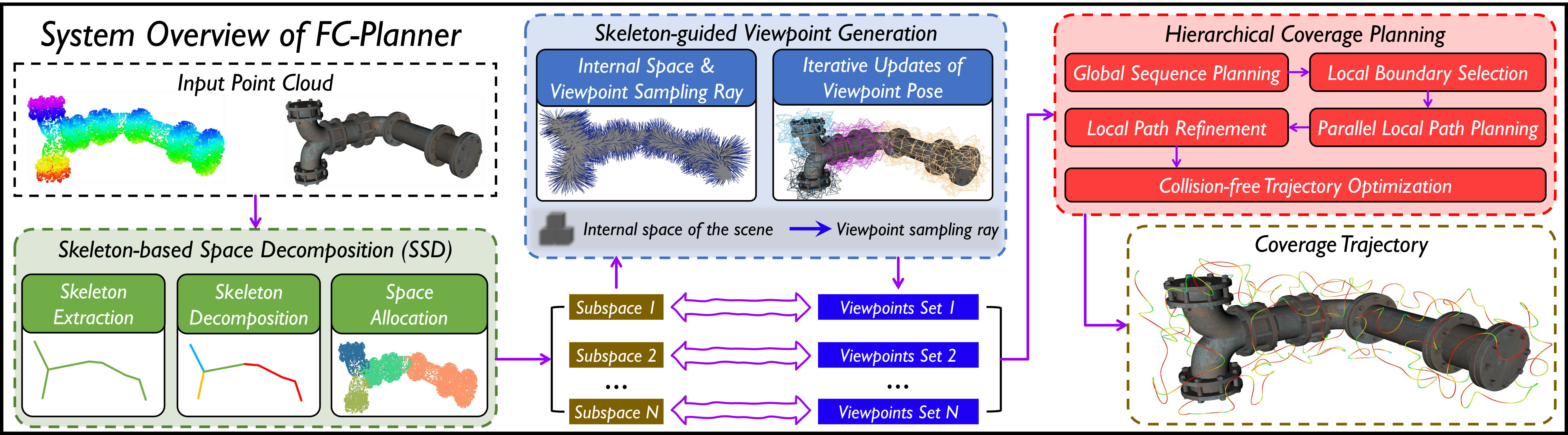}    
      \vspace{-0.5cm}
	\end{center}
   \caption{\label{fig:overview} The overview of the proposed skeleton-guided planning framework for fast aerial coverage in complex 3D scenes.}
   \vspace{-0.7cm}
\end{figure*}

% system introduce 
To tackle the above limitations, we propose \textbf{FC-Planner}, a skeleton-guided planning framework tailored for fast coverage of large and complex 3D scenes.
Central to our method is the skeleton-based space decomposition (SSD), which efficiently extracts the skeleton of target scenes.
The skeleton provides useful guidance in two aspects.
Firstly, it is located inside the target scene, enabling effortless differentiation between internal and external spaces, eliminating the necessity for the pre-processing of free space.
Secondly, the skeleton decomposes the complex scene into subspaces characterized by simple geometry. 
Then, viewpoint sampling rays originated from the skeleton of each subspace are employed to generate specialized viewpoints, thus avoiding redundant sampling of viewpoints and mismatches between viewpoints and their corresponding subspaces.
Further, a gravitation-like method is designed to screen out a minimal set of informative viewpoints.
Leveraging the straightforward geometry of each subspace and its dedicated set of viewpoints, our approach effectively divides the entire coverage path planning problem into parallelizable sub-problems.
This, in conjunction with our carefully designed global and local planning procedures, results in the generation of high-quality coverage paths and high computational efficiency.

% benchmark and contributions
We compare our proposed method with state-of-the-art works in simulation.
Results show that our method achieves significantly higher computational efficiency (over 10 times faster) and more complete coverage with shorter paths.
To further validate \textbf{FC-Planner}, we conduct real-world coverage tests in a challenging scene.
Both the simulation and real-world experiments demonstrate the superior system simplicity and performance of our method compared to state-of-the-art ones.
The contributions of this paper are as follows:

1) An SSD that efficiently extracts the skeleton of complex 3D scenes and guides the coverage planning. 

2) An efficient and specialized viewpoint sampling guided by the skeleton, and a gravitation-like model that efficiently screens out the minimal set of informative viewpoints. 

3) A hierarchical coverage planner, which divides the entire planning problem into parallelizable sub-problems utilizing SSD. Along with the carefully designed global and local planning modules, it ensures high path quality and computational efficiency.

4) Extensive experiments validate the proposed method. The source code of our implementation will be made public.

\section{Related Work}
\label{sec:related}

The coverage path planning problem (CPP) in 3D scenes has been an active research topic for UAVs with a wide range of practical applications. 
Existing methods typically adopt a sampling-based framework to solve such CPP in two steps, namely viewpoint generation and path planning.

\noindent\textbf{Viewpoint Generation:}
It refers to the process of obtaining a set of viewpoints to fulfill the coverage need of a target scene. 
Existing methods required pre-processing the scene to identify free space, from which safe and valid viewpoints are then sampled.
Some of these methods \cite{bircher2015structural, jing2016sampling, tong2022uav, almadhoun2018coverage, jing2019coverage, roberts2017submodular} utilized triangles of the mesh to sample viewpoints.
Alternatively, other works \cite{cao2020hierarchical, huang2023bim, claro2023energy} chose the form of point clouds.
Our approach falls into the latter category.
However, both types often sampled a large number of viewpoint candidates and then iteratively selected a subset of viewpoints to achieve complete coverage.
This strategy resulted in vast time consumption for selection and made it challenging to find the fewest viewpoints needed.

In contrast, our method avoids pre-processing by utilizing the skeleton to determine free space and then samples viewpoints around the scene.
To efficiently find a minimal set of viewpoints for full coverage, we develop an efficient strategy that iteratively updates the poses of viewpoints using a gravitation-like model without time-consuming selection.

\noindent\textbf{Path Planning:}
The objective of this step is to plan the shortest collision-free path while comprehensively covering the scene.
Previous approaches \cite{bircher2015structural, jing2016sampling, almadhoun2018coverage, jing2019coverage, huang2023bim, roberts2017submodular, tong2022uav} formulated this problem as some concrete forms of the COP, \textit{e.g.}, Traveling Salesman Problem (TSP) \cite{dantzig1954solution}, Submodular Orienteering Problem (SOP) \cite{chekuri2005recursive}, etc.
As such problems are known to be NP-hard, the computation time increases drastically with the complexity of the scene. 
To address this issue, HCPP \cite{cao2020hierarchical} proposed an approach that divides the scene using an Octree.
Each leaf in the Octree contains a maximum of $N$ viewpoints, defining a subspace.
Then, the planning problem is transformed into solving local TSPs for each subspace.
However, having viewpoints within a single subspace does not guarantee complete coverage of that subspace. 
Additional viewpoints from other subspaces may be required to achieve full coverage.
Thus, the covered area for each subspace cannot be pre-allocated, and the covered goals for the next subspace can only be updated once the planning for the previous subspace is completed.
This poses a challenge in achieving parallel planning for different subspaces.

Instead, our SSD decomposes the scene into several simple subspaces guided by the skeleton.
This ensures that viewpoints within each subspace avoid mismatches with this subspace.
As a result, parallel path generation for each subspace becomes possible, leading to significant savings in computation time.

% System Overview

\section{System Overview}
\label{sys}

The proposed framework takes the point cloud of the target scene as input, which is widely adopted in various applications.
As shown in Fig.\ref{fig:overview}, it consists of SSD (Sect.\ref{sec:ssd}), efficient viewpoint generation (Sect.\ref{sec:evg}), and hierarchical coverage planner (Sect.\ref{sec:hcp}).
SSD extracts the skeleton of the point cloud and decomposes the cloud into subspaces with simple geometry (Sect.\ref{sec:ssd}).
Guided by the skeleton, we determine the internal space and viewpoint sampling rays to sample safe initial viewpoints without pre-processing.
Our method then minimizes the number of viewpoints by updating their poses using a gravitation-like model (Sect.\ref{sec:evg}).
Afterwards, the hierarchical planner finds a global visiting sequence of subspaces, optimizes local paths for each subspace in parallel, refines the junction parts between local paths, and generates a trajectory to realize complete coverage (Sect.\ref{sec:ssd}).

% \begin{figure}[t]
% 	\begin{center}
%       \includegraphics[width=0.80\columnwidth]{rosa.png}     
%       \vspace{-0.5cm}
% 	\end{center}
%    \caption{\label{fig:rosa} Illustration of finding ROSA point.
%    }
%    \vspace{-1.0cm}
% \end{figure}

\section{Skeleton-based Space Decomposition}
% Space Decomposition
\label{sec:ssd}

In this section, we present our skeleton-based space decomposition in three steps.
1) Compute the skeleton composed of generalized rotational symmetry axis (ROSA) points \cite{tagliasacchi2009curve} given the point cloud of the target scene (Sect.\ref{subs:ske_ext}).
2) Decompose the skeleton into several simple shapes, termed branches (Sect.\ref{subs:ske_decomp}).
3) Allocate the point cloud into each branch, forming the corresponding subspace (Sect.\ref{subs:sp_alloc}).
 
\subsection{Skeleton Extraction}
\label{subs:ske_ext}

To begin, we obtain the oriented point cloud $\mathcal{P}_O$ from the input point cloud by normal estimation \cite{rusu20113d}.
Inspired by \cite{tagliasacchi2009curve}, we represent the skeleton using a set of ROSA points, each of which consists of a position and an orientation denoted as $r$=$({\rm \textbf{x}}, {\rm \textbf{v}})$.
To accelerate the calculation process, we downsample $\mathcal{P}_O$ as $\mathcal{P}_D$ and compute its ROSA point $r_p$=$({\rm \textbf{x}}_p, {\rm \textbf{v}}_p)$ of each point $p\in\mathcal{P}_D$.
For each $p$, its corresponding ROSA point operates on the $p$'s neighborhood $\mathcal{N}_{p}$ within the plane constructed by ${\rm \textbf{v}_p}$ and $p$. 
ROSA point exhibits high rotational symmetry about $\mathcal{N}_{p}$.
This property necessitates that the orientation ${\rm \textbf{v}_p}$ minimizes the variance of the angles between ${\rm \textbf{v}}_p$ and normals in $\mathcal{N}_{p}$, while the position ${\rm \textbf{x}_p}$ minimizes the sum of squared distances to the line extensions of normals in $\mathcal{N}_{p}$.
Firstly, given an initial orientation ${\rm \textbf{v}}_p^0$, we iteratively optimize ${\rm \textbf{v}}_p$ by solving the below quadratic programming problem:
\begin{equation}
   {\rm \textbf{v}}_p^{i+1} = \mathop{\arg\min}\limits_{{\rm \textbf{v}}_p\in\mathbb{R}^3, ||{\rm \textbf{v}}_p||_2=1} {{\rm \textbf{v}}_p^{i}}^\mathsf{T}\Sigma_{p}^{i}{\rm \textbf{v}}_p^{i},
\end{equation}
where ${\rm \textbf{v}}_p^{i}$ denotes the orientation at the $i$-th iteration.
$\Sigma_{p}^{i}$ is the covariance matrix of all point normals in $p$'s neighborhood $\mathcal{N}_p^i$.
After finding orientation of ROSA point, we calculate ${\rm \textbf{x}}_p$ using the following procedure:
\begin{equation}
   \mathop{\arg\min}\limits_{{\rm \textbf{x}}_p\in\mathbb{R}^3}\sum\limits_{p_k \in \mathcal{N}_p}||({\rm \textbf{x}}_p-p_k) \times {\rm \textbf{n}}(p_k)||_2^2,
\end{equation}
where ${\rm \textbf{n}}(p_k)$ denotes point normal of $p_k \in \mathcal{N}_p$.
Lastly, we apply 1D moving least square \cite{lee2000curve} on all ROSA points to generate the skeleton.
The skeleton is stored in an undirected graph $\mathcal{G} = (V_s, E_s)$, where vertices $V_s$ contain all ROSA points while edges $E_s$ represent their connections.

% \vspace{-0.3cm}
\subsection{Skeleton Decomposition}
\label{subs:ske_decomp}

As outlined in Algo.\ref{alg1}, we decompose the extracted skeleton into several branches, each of which comprises a set of edges in $\mathcal{G}$.
These branches are all contained in the set of branches denoted by $\mathcal{B}$.
In line 2, we identify joints $\mathcal{J}$ on $\mathcal{G}$, which are defined as vertices with their degree greater than two.
Due to simple geometry requirements, a branch typically starts with a joint and ends with a joint or a leaf vertex (degree equals one).
To obtain branches satisfying the above needs, we develop a depth-first search (DFS)-like procedure \cite{tarjan1972depth} in lines 3-14.
Further, to ensure that each branch is as simple and easy to cover as possible, we require that edges within one branch should have similar directions.
Thus, in lines 16-26 we decompose the branches with large direction changes into simpler ones.

\begin{algorithm}[t]
   \label{alg1}
   \footnotesize
   \caption{Skeleton Decomposition} 
   \KwIn{Skeleton graph $\mathcal{G} = (V_s, E_s)$, Angle variance maximum $\delta$}
   \KwOut{Branches set $\mathcal{B}$}
   \textbf{Define:} Joints set $\mathcal{J}$, Branch $\mathcal{T}_b$, Current vertex $v_c$, Next vertex $v_n$ \\
   % \For{$v\in V_s$}
   % {\If{deg(v) $>$ 2}{$\mathcal{J}$.Add(v);}}
   $\mathcal{J} \leftarrow {\rm FindJoint}(V_s)$; \\
   \For{$v_j\in\mathcal{J}$}
   {
      \For{$v_{adj}\in v_j$.$\rm AdjacentVertices$}
      {
         $\mathcal{T}_b$.clear(); $v_c=v_j$, $v_n=v_{adj}$; \\
         $\mathcal{T}_b$.AddEdge($v_{c}v_{n}$); \\
         \While{$deg(v_n)=2$}
         {
            $\mathcal{Q} \leftarrow v_n.\rm AdjacentVertices$.remove($v_c$); \\   
            $v_c=v_n$, $v_n=\,$$\mathcal{Q}$.first(); \\
            $\mathcal{T}_b$.AddEdge($v_{c}v_{n}$); \\
         }
         $\mathcal{B}$.Add($\mathcal{T}_b$); \\
      }
   }
   \textbf{Define:} Temporary branches set $\mathcal{B}_T$, Reference direction angle $\sigma$ \\
   \For{$b\in\mathcal{B}$}
   {
      $\mathcal{T}_b$.clear(); $\mathcal{T}_b$.AddEdge($b[0]$); $\sigma={\rm CalculateAngle}(b[0])$; \\
      \For{$i \in [1,b$.size()$)$}
      {
         \eIf{$||{\rm CalculateAngle}(b[i])-\sigma||<\delta$}
         {
            $\mathcal{T}_b$.AddEdge($b[i]$);
         }
         {
            $\mathcal{B}_T$.Add($\mathcal{T}_b$); $\mathcal{T}_b$.clear(); \\ 
            $\mathcal{T}_b$.AddEdge($b[i]$); $\sigma={\rm CalculateAngle}(b[i])$; \\
         }
      } 
   }
   $\mathcal{B}=\mathcal{B}_T$.
\end{algorithm}
\setlength{\textfloatsep}{-0.5cm}

\vspace{-0.4cm}
\subsection{Space Allocation}
\label{subs:sp_alloc}
\vspace{-0.2cm}

Given the decomposed skeleton, we aim to allocate the points in $\mathcal{P}_O$ into multiple subspaces, each of which corresponds to a branch in $\mathcal{B}$.
We discretize edges in those branches into several oriented points, whose orientation is the direction of the edge it is on.
The planes determined by these oriented points are denoted as ${\Gamma}$.
Then, for each plane, we query points within this plane from $\mathcal{P}_O$.
If one point $p\in\mathcal{P}_O$ is located on more than one plane, $p$ will be allocated to the plane whose oriented point is nearest to it.
After that, the points queried by planes (named allocated points $\mathcal{A}$) from the same branch are included in a subspace.
All subspaces are contained in $\mathcal{S}=\{s_1,..., s_N\}$.

\vspace{-0.4cm}

\section{Skeleton-guided Viewpoint Generation}
% Viewpoint Generation
\label{sec:evg}
\vspace{-0.3cm}

This section presents the efficient generation of viewpoints for full coverage. 
The skeleton allows for the identification of internal space that ensures the safety of viewpoints without pre-processing. 
Additionally, the skeleton spawns a set of viewpoint sampling rays, guiding the production of viewpoints (Sect.\ref{subs:i_and_vr}). 
To efficiently find a minimal set of viewpoints, we propose a gravitation-like model that iteratively adjusts the pose of viewpoints and removes redundant ones, screening out a more representative subset (Sect.\ref{subs:obvs}). This process is further accelerated by bidirectional ray casting (BiRC) that allows for faster visibility checks (Sect.\ref{subs:birc}).

\subsection{Internal Space and Viewpoint Sampling Ray}
\label{subs:i_and_vr}

\begin{figure}[t]
	\begin{center}
      \vspace{0.2cm}
      \includegraphics[width=0.9\columnwidth]{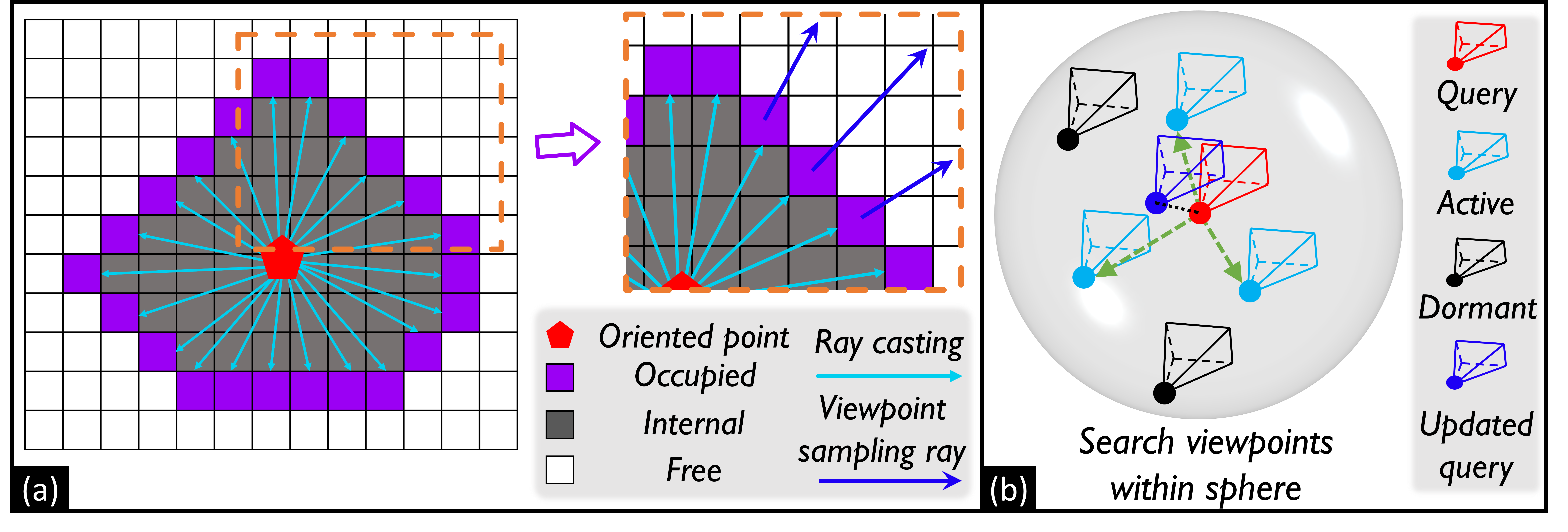}     
      \vspace{-0.5cm}
	\end{center}
   \caption{\label{fig:vg_2} (a) Generation of internal space and viewpoint sampling rays (Sect.\ref{subs:i_and_vr}). (b) Gravitation-like model used to update the pose of query viewpoint (Sect.\ref{subs:obvs}).
   }
   \vspace{0.1cm}
\end{figure}

The internal space is represented using a volumetric map whose all voxels are initialized as free.
The voxels that contain $\mathcal{P}_O$ are updated as occupied.
For each plane $\gamma_i \in \Gamma$ (Sect.\ref{subs:sp_alloc}), we perform ray casting from its oriented point $o_i$ (located on the skeleton) to each of its allocated points $\mathcal{A}_i$.
Fig.\ref{fig:vg_2}(a) shows that voxels along the rays before crossing the occupied ones are labeled as internal.
The rays continue to extend into the free space, and we define the parts of these rays starting from the occupied voxels as viewpoint sampling rays.
Safe and informative viewpoints are sampled along them, reducing the number of redundantly sampled viewpoints.
These specialized viewpoints also provide better coverage since the skeleton guides them all pointing towards the surface.
Moreover, the viewpoint sampling rays that cast from $o_i$ belong to the subspace that corresponds to $o_i$.

\vspace{-0.2cm}

\subsection{Bidirectional Ray Casting}
\label{subs:birc}

\vspace{-0.1cm}

Ray casting is widely used for the visibility check of viewpoints.
However, this approach traverses voxels unidirectionally, commencing from one end of the ray. 
It may not efficiently identify occlusions when an occupied voxel is located near the other end of the ray. 
To address this issue, we implement bidirectional ray casting (BiRC), wherein voxels are traversed from both ends toward the midpoint of the ray. 
This approach stops earlier and eliminates redundant checks in the face of the aforementioned situation.

\vspace{-0.2cm}

\subsection{Iterative Updates of Viewpoint Pose}
\label{subs:obvs}

\vspace{-0.1cm}

Our method employs 5-DoF viewpoints, denoted as $\textbf{vp}=[{\rm \textbf{p}},\theta,\phi, id]$, where ${\rm \textbf{p}}$ is the 3D position of the sensor, $\theta,\phi$ respectively represent pitch and yaw angles.
The parameter $id$ is the subspace to which the viewpoint belongs.
The start and direction of viewpoint sampling ray $r_{vs}$ is respectively described as $\textbf{sr}=[x_{sr}, y_{sr}, z_{sr}]$ and $\textbf{dr}=[nx_{dr},ny_{dr},nz_{dr}]$.
We sample one viewpoint along each $r_{vs}$ at distance $D$ away, the $id$ of which refers to the subspace $r_{vs}$ belongs to.
The position, pitch, and yaw of the sampled viewpoint are determined as follows:
\begin{align}
   & {\rm \textbf{p}} = \textbf{sr} + D*\textbf{dr}, \theta = {\rm arcsin}(-nz_{dr}/||\textbf{dr}||_2), \nonumber \\
   & \phi = {\rm arctan}(-ny_{dr}/-nx_{dr}).
\end{align}
These sampled viewpoints are named as initial viewpoints $\mathcal{VP}_{ini}$.
The first step is to determine voxels covered by each viewpoint in $\mathcal{VP}_{ini}$ using the proposed BiRC.
Next, to reduce the number of viewpoints, we assign each voxel to the viewpoint that covers the most voxels if it is observed by more than two viewpoints.
We then remove the viewpoints that are not assigned any voxels from $\mathcal{VP}_{ini}$.
Afterwards, we construct a kd tree \cite{zhou2008real} $T_{ini}$ for $\mathcal{VP}_{ini}$ and define a binary state (active or dormant) for each viewpoint, initially setting all viewpoints as active.
To ensure that each viewpoint covers as many voxels as possible, we merge the viewpoints with fewer voxels into the viewpoints with more voxels using a gravitation-like model.
Specifically, commencing with the viewpoint that covers the maximum number of voxels and proceeding to the one that covers the minimum, each viewpoint (denoted as $\textbf{vp}_q$) queries its neighborhood $\mathcal{VP}_q$ from $T_{ini}$ using a radius search.
The radius $r_{q}$ is determined by maximal visible distance $d_v$ and the Field of View (FoV) $[f_h, f_w]$:
\begin{equation}
   r_{q} = d_v*{\rm tan}({\rm min}(f_h,f_w)/2).
\end{equation}
Fig.\ref{fig:vg_2}(b) illustrates how we update the pose of the query viewpoint $\textbf{vp}_q$ using all active viewpoints $\mathcal{VP}_a$ in $\mathcal{VP}_q$ through a gravitation-like model:
\begin{equation}
   \overline{{\rm \textbf{p}}_q} = {\rm \textbf{p}}_q + \sum\limits_{\textbf{vp}_a\in\mathcal{VP}_a}\frac{c_a}{c_q}({\rm \textbf{p}}_a-{\rm \textbf{p}}_q), \; {\rm s.t.} \; c_a < c_q,
\end{equation}
where $c_q$ and $c_a$ denote the number of voxels covered by $\textbf{vp}_q$ and each viewpoint in $\mathcal{VP}_a$, respectively.
The updated position of $\textbf{vp}_q$ is represented as $\overline{{\rm \textbf{p}}_q}$.
Similarly, we obtain the updated pitch $\overline{\theta_q}$ and yaw $\overline{\phi_q}$.
They are further adjusted to distribute voxels covered by $\textbf{vp}_q$ around the center of FoV.
The subspace to which $\textbf{vp}_q$ belongs is the same as that of the nearest viewpoint to $\textbf{vp}_q$ in $\mathcal{VP}_{ini}$.
Next, we set the state of the viewpoints used to update $\textbf{vp}_q$ in $\mathcal{VP}_a$ as dormant.
After that, we identify uncovered voxels by performing the first step for all active viewpoints and then sample new viewpoints to cover them, denoted as $\mathcal{VP}_{unc}$.
Lastly, we repeat the above steps for $\mathcal{VP}_{unc}$ achieving complete coverage.
After such iterative updates for the poses of viewpoints, all remaining active viewpoints are assigned to each subspace in $\mathcal{S}$ according to their $id$, with each subset of viewpoints represented as $\mathcal{V}=\{\mathcal{VP}_1, ..., \mathcal{VP}_N\}$.

\begin{algorithm}[t]
   \label{alg2}
   \footnotesize
   \caption{Local Path Refinement} 
   \KwIn{Coverage path $\mathcal{P}_\mathcal{C}$, branches set $\mathcal{B}$, junction radius $r_{jc}$}
   \KwOut{Refined coverage path $\mathcal{P}_\mathcal{R}$}
   $\mathcal{Z} \leftarrow {\rm FindJunctionVertex}(\mathcal{B})$; $T_\mathcal{C} \leftarrow {\rm BuildKdTree}(\mathcal{P}_\mathcal{C})$; \\
   \For{$z\in\mathcal{Z}$}
   {
      $V_{jc} \leftarrow T_\mathcal{C}.{\rm RadiusSearch}(z, r_{jc})$; \\
      \For{$i\in[1,K]$}
      {
         $v_1 \leftarrow {\rm RandomPick}(V_{jc})$; $V_{temp} \leftarrow V_{jc}$.remove$(v_1)$;\\
         $v_2 \leftarrow {\rm RandomNeighbor}(v_1)$;\\
         \eIf{$v_2==v_1.Suc$}
         {
            $V_{temp} \leftarrow V_{temp}$.remove$(v_2 \, \& \, v_2.Suc)$;
         }
         {
            $V_{temp} \leftarrow V_{temp}$.remove$(v_2 \, \& \, v_2.Pred)$;
         }
         $v_3 \leftarrow {\rm RandomPick}(V_{temp})$; \\ 
         $\mathcal{P}_\mathcal{R} \leftarrow {\rm Make2Opt}(v_1, v_2, v_3)$;
      }
   }
\end{algorithm}
\setlength{\textfloatsep}{-0.5cm}

\vspace{-0.1cm}

\section{Hierarchical Coverage Planning}
% Hierarchical Planning
\label{sec:hcp}

Given the above viewpoints, this stage endeavors to find the shortest collision-free path traversing them.
Supported by SSD, we decompose large planning problem into several independent and parallelizable sub-problems in five steps.

\begin{figure}[t]
	\begin{center}
      \vspace{0.2cm}
      \includegraphics[width=0.9\columnwidth]{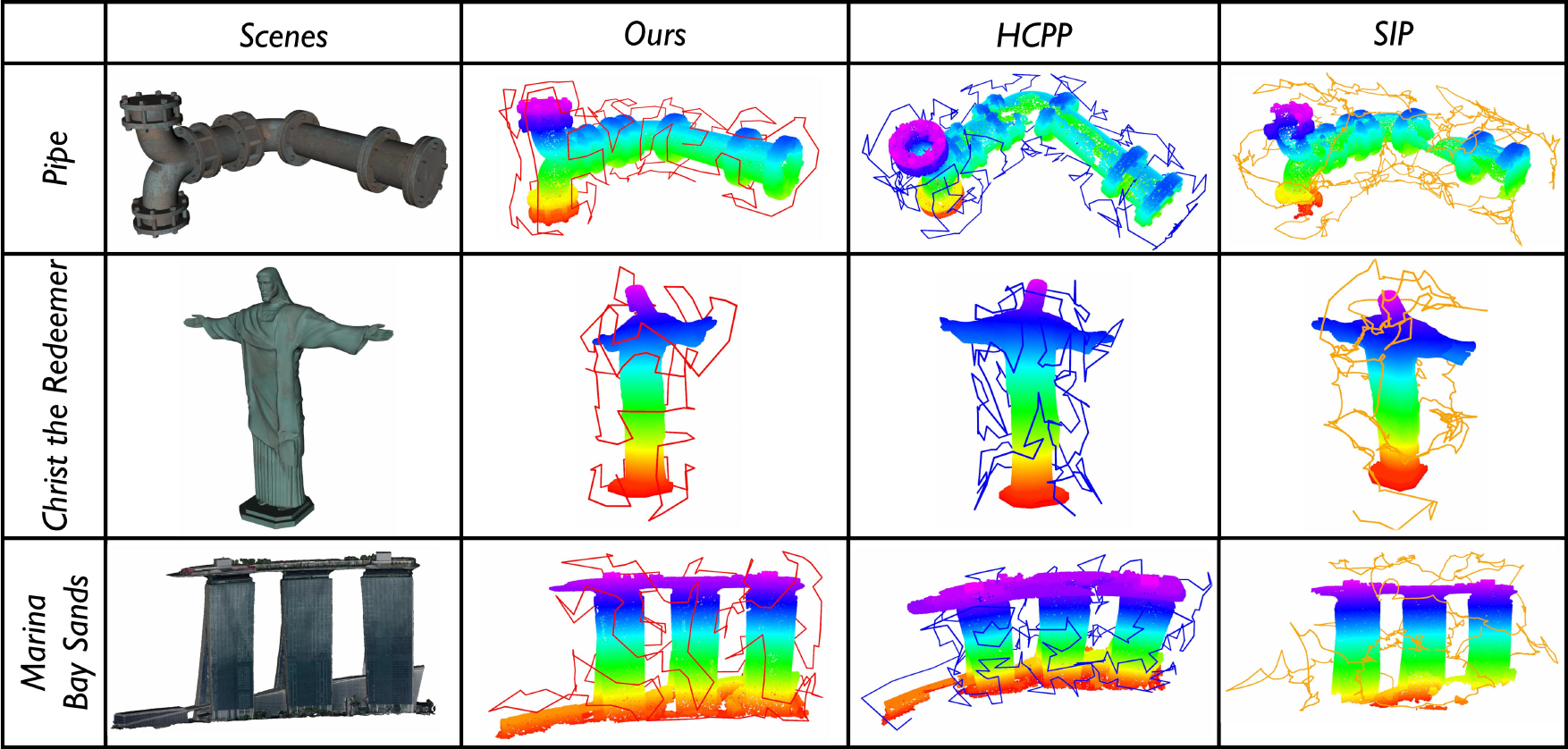}     
      \vspace{-0.5cm}
	\end{center}
   \caption{\label{fig:qua} Comparisons on coverage situation and generated path of the proposed method, SIP \cite{bircher2015structural}, and HCPP \cite{cao2020hierarchical} in three complex scenarios.
   }
   \vspace{0.5cm}
\end{figure}

\noindent\textbf{Global Sequence Planning.}
It seeks to solve an optimal tour visiting all subspaces starting from UAV's current pose, which is defined as an Asymmetric Traveling Salesman Problem (ATSP) \cite{meng2017two}.
We first calculate the centroid of viewpoints in each subspace and then construct a Euclidean distance matrix $\mathcal{M}_G$, between these centroids and UAV's current pose.
Lin-Kernighan-Helsgaun (LKH) solver \cite{helsgaun2000effective} takes $\mathcal{M}_G$ as input to obtain global sequence $[g_1,...,g_N]$.

\noindent\textbf{Local Boundary Selection.}
We expect to select suitable start and end viewpoints for each subspace, ensuring that the results of local planning are compatible with the global sequence.
According to global sequence, we sort all centroids as $Seq_C = [{\rm \textbf{k}}_0, ..., {\rm \textbf{k}}_N]$, where ${\rm \textbf{k}}_0$ is UAV's current pose and others are sorted centroids.
Based on $Seq_C$, the start $\textbf{vp}_{start}^{g_i}$ and end $\textbf{vp}_{end}^{g_i}$ of $i$-th visited subspace are obtained as:
\begin{equation}
   \begin{split}
      & \textbf{vp}_{start}^{g_i} = \mathop{\arg\min}\limits_{\textbf{vp}_{g_i}\in \mathcal{VP}_{g_i}}||{\rm \textbf{p}}_{g_i}-{\rm \textbf{k}}_{i-1}||_2^2+||{\rm \textbf{p}}_{g_i}-{\rm \textbf{k}}_{i}||_2^2, \\
      & \textbf{vp}_{end}^{g_i} = \mathop{\arg\min}\limits_{\textbf{vp}_{g_i}\in \mathcal{VP}_{g_i}}||{\rm \textbf{p}}_{g_i}-{\rm \textbf{k}}_{i}||_2^2+||{\rm \textbf{p}}_{g_i}-{\rm \textbf{k}}_{i+1}||_2^2.
   \end{split}
\end{equation}
In particular, the last visited subspace will not have an end viewpoint since there is no subsequent subspace.

\noindent\textbf{Parallel Local Path Planning.}
Once determining boundary viewpoints, we can individually plan local paths for each subspace.
We formulate this problem as a variant of TSP that adds a constraint of given start and end.
Suppose a subspace has $R$ viewpoints, cost matrix $\mathcal{M}_L$ for this TSP follows:
\begin{align}
   & \mathcal{M}_L(i,j) =
   \begin{cases}
      0,&i==j \,or\, j=0 \nonumber \\
      +\infty,&i=R-1 \,and\, j\in\{1,...,R-2\} \nonumber \\
      c_{\textbf{vp}_i,\textbf{vp}_j},&others \\
   \end{cases}, \\
   & c_{\textbf{vp}_i,\textbf{vp}_j} = {\rm max}({\rm max}(\frac{{\rm L}({\rm \textbf{p}}_i, {\rm \textbf{p}}_j)}{{\rm v}_{max}},\frac{{\rm Ang}(\theta_i,\theta_j)}{\omega_{max}}), \frac{{\rm Ang}(\phi_i,\phi_j)}{\omega_{max}}), \nonumber \\
   & {\rm Ang}({\rm a}_1, {\rm a}_2) = {\rm min}(|{\rm a}_1-{\rm a}_2|,2\pi-|{\rm a}_1-{\rm a}_2|),
\end{align}
where ${\rm L}(\cdot)$ is the path length between $v_i$ and $v_j$ searched by $A^*$ algorithm to warrant collision-free path, ${\rm v}_{max}$ and $\omega_{max}$ are the limits of velocity and angular rate of pitch and yaw.
For the last visited subspace, all $+\infty$ entries in $\mathcal{M}_L$ are set as $c_{\textbf{vp}_i,\textbf{vp}_j}$.
For each $\mathcal{VP}_i\in\mathcal{V}$, we use multi-thread programming to parallelly produce local path that covers its corresponding subspace.
Let there be a total of $m\in\mathbb{R}^+$ viewpoints and the maximal number of viewpoints in all sub-problems is $e\in\mathbb{R}^+$, $m \gg e$.
Since the time complexity of solving a TSP using the LKH solver is $\mathcal{O}(n^{2.2})$ \cite{papadimitriou1992complexity}, solving for the TSP of all viewpoints without problem decomposition takes $\mathcal{O}(m^{2.2})$ time.
Thanks to our parallel planning, the time complexity of solving for all sub-problems is reduced to $\mathcal{O}(e^{2.2})$.
Hence, our method effectively lowers the time complexity and improves computational efficiency.
We combine all local paths in the order of global sequence to obtain the entire coverage path $\mathcal{P}_\mathcal{C}$.

\begin{table}[t]
   \scriptsize
   \renewcommand\arraystretch{1.4}
   \tabcolsep=0.4mm
   \centering
   \vspace{0.2cm}
   \caption{Coverage Planning Comparisons in Three Scenarios.  \label{tab:benchmark}}
   \vspace{-0.2cm}
   \begin{tabular}{c|l|cccccc} 
   \hline
                     Scenario & Method  & Pre-process & \begin{tabular}[c]{@{}c@{}}Comp. \\Time (ms) \end{tabular} & \begin{tabular}[c]{@{}c@{}}Viewpoint \\Number\end{tabular} & \begin{tabular}[c]{@{}c@{}} Path \\Length (m)\end{tabular}  & \begin{tabular}[c]{@{}c@{}}Exec. \\Time (s)\end{tabular} & \begin{tabular}[c]{@{}c@{}}Coverage \\ Rate (\%)\end{tabular} \\
   \hline
   \hline
   \multirow{3}{*}{\rotatebox{90}{\textit{Pipe}}}     & SIP \cite{bircher2015structural}      & $\checkmark$ & 235012.1                                                          & 3078                                                        & 2754.1 & 3938.8 & 69.2                                                          \\ 
   \cline{2-8}
                             & HCPP \cite{cao2020hierarchical}        & $\checkmark$ &   16934.2                                                       & 446                                                          & 1874.3 &1670.9 &87.3                                                           \\ 
   \cline{2-8}
                             & Ours                    & \textcolor{red}{\ding{56}}  & \textbf{1614.8}                                                     & \textbf{210}                                                        & \textbf{1440.4} & \textbf{1033.4} & \textbf{97.5}                                                          \\ 
   \hline
   \multirow{3}{*}{\rotatebox{90}{\begin{tabular}[c]{@{}c@{}}\textit{Christ the} \\\textit{Redeemer$\,$} \end{tabular}}} & SIP \cite{bircher2015structural}       & $\checkmark$  & 62110.9                                                         & 787                                                         & 700.5 & 1104.8 & 85.9                                                          \\ 
   \cline{2-8}
                             & HCPP \cite{cao2020hierarchical}     & $\checkmark$  & 5485.7                                                         & 228                                                          & 635.0 & 642.3 &97.4                                                           \\ 
   \cline{2-8}
                             & Ours                 & \textcolor{red}{\ding{56}}    & \textbf{506.7}                                                       & \textbf{111}                                                         & \textbf{530.4} & \textbf{386.9} & \textbf{99.7}                                                           \\
   \hline
   \multirow{3}{*}{\rotatebox{90}{\begin{tabular}[c]{@{}c@{}}\textit{Marina} \\\textit{Bay Sands$\,$} \end{tabular}}} & SIP \cite{bircher2015structural}       & $\checkmark$  & 131454.8                                                         & 1770                                                         & 1635.5 & 2789.6 & 68.5                                                           \\ 
   \cline{2-8}
                             & HCPP \cite{cao2020hierarchical}     & $\checkmark$  & 12639.4                                                         & 369                                                          & 1698.9 & 1613.9 & 89.3                                                           \\ 
   \cline{2-8}
                             & Ours                 & \textcolor{red}{\ding{56}}    & \textbf{954.3}                                                       & \textbf{185}                                                         & \textbf{1396.2} & \textbf{1178.5} & \textbf{94.0}                                                          \\
   \hline
   \end{tabular}
   \vspace{0.1cm}
\end{table}

\noindent\textbf{Local Path Refinement.}
Since the coverage path in the previous step is optimized within each subspace, it cannot guarantee global optimality thus it may produce unnecessary detours.
We have observed that these detours typically occur near the junctions between local paths.
Thus, we propose a local refinement strategy to further improve the entire coverage path on junctions, as depicted in Algo.\ref{alg2}.
A junction vertex is defined as a vertex where two branches intersect in line 1.
Lines 2-16 show the iterative refinement procedure using a local search in two steps, which continues $K$ iterations.
It first chooses the parts from the whole coverage path that are near the junctions.
Then, each iteration randomly selects three viewpoints from these parts and swaps them using 2-opt \cite{mcgovern20042} if this swap can shorten the coverage path.

\noindent\textbf{Collision-free Trajectory Optimization.}
It aims to convert refined coverage path $\mathcal{P}_\mathcal{R}=\{v_{\mathcal{R}}^0, ..., v_{\mathcal{R}}^M\}$ to a smooth, dynamically feasible, safe, and minimum-time trajectory passing through all viewpoints.
Specifically, we divide $\mathcal{P}_\mathcal{R}$ into $M$ trajectory pieces and then generate 3D Safe Flight Corridors (SFCs), where each convex flight corridor is assigned with a consecutive set of trajectory pieces.
For position trajectory, SFCs are used to ensure safety:
\begin{equation}
   \vspace{-0.1cm}
   tp_i(t) \in CP(i), \forall t\in[0,T_i], \forall 1 \leq i \leq M,
   \vspace{-0.05cm}
\end{equation}
where $tp_i$ is the $i$-th trajectory piece with its duration $T_i$ and $CP(i)$ denotes the assigned convex flight corridor for $tp_i$ in SFCs.
To avoid visual blur caused by aggressive flight as well as ensuring dynamic feasibility, the velocity, acceleration and jerk limits are also considered:
\begin{align}
   \vspace{-0.1cm}
   \label{eqdynamic}
   \begin{split}
      & ||\dot{tp_i(t)}||_2^2 \leq {\rm v}_{max}^2, \forall t\in[0,T_i], \forall 1 \leq i \leq M, \\
      & ||\ddot{tp_i(t)}||_2^2 \leq {\rm a}_{max}^2, \forall t\in[0,T_i], \forall 1 \leq i \leq M, \\
      & ||\dddot{tp_i(t)}||_2^2 \leq {\rm j}_{max}^2, \forall t\in[0,T_i], \forall 1 \leq i \leq M, \\
      & {\rm {s.t.}} \; tp_i(0) = {\rm \textbf{p}}_\mathcal{R}^{i-1}, tp_i(T_i) = {\rm \textbf{p}}_\mathcal{R}^{i},
   \end{split}
   \vspace{-0.1cm}
\end{align}
where ${\rm v}_{max}$, ${\rm a}_{max}$, and ${\rm j}_{max}$ are dynamic limits.
We also generate trajectories for pitch and yaw angles, where the maximum angular velocity is limited similarly to Eq.\ref{eqdynamic}.
We utilize MINCO \cite{wang2022geometrically} to optimize the time allocation of coverage trajectory while satisfying the above constraints.

\vspace{-0.2cm}
\section{Experiments}
\label{sec:exp}

\begin{figure}[t]
	\begin{center}
      \vspace{0.2cm}
      \includegraphics[width=0.90\columnwidth]{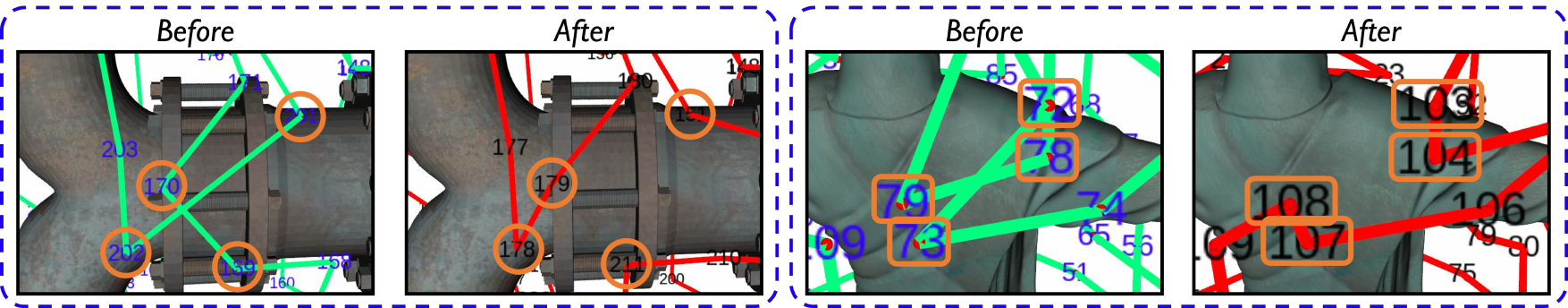}     
      \vspace{-0.5cm}
	\end{center}
   \caption{\label{fig:refine} Comparisons of results before and after using local path refinement.
   }
   \vspace{-0.4cm}
\end{figure}

\subsection{Implementation Details}

\vspace{-0.1cm}

During the skeleton extraction step, we normalize the input point cloud within a unit sphere, which enhances its robustness to variations in scene scale
We set $\delta=45^{\circ}$ in Algo.\ref{alg1}, the number of iterations in local path refinement $K=10000$ since each iteration only consumes around 0.01ms.
In all experiments, we choose a 2-axis gimbal camera as the sensor equipped with the UAV.
A geometric controller \cite{lee2010} is used for tracking control of (${\rm \textbf{p}},\theta,\phi$) trajectory.
All modules run on an Intel Core i9-13900K CPU.

\vspace{-0.25cm}

\subsection{Benchmark Comparisons and Analysis}

\vspace{-0.1cm}

To validate our proposed framework, we benchmark it in simulation including three large and complex scenes, \textit{Pipe} (\textbf{73}$\times$\textbf{66}$\times$\textbf{41} m$^3$), \textit{Christ the Redeemer} (\textbf{10}$\times$\textbf{36}$\times$\textbf{38} m$^3$), and \textit{Marina Bay Sands} (\textbf{100}$\times$\textbf{25}$\times$\textbf{49} m$^3$).
Two state-of-the-art methods are compared: SIP \cite{bircher2015structural} and HCPP \cite{cao2020hierarchical}.
There is no open source code for HCPP \cite{cao2020hierarchical}, so we use our implementation.
In simulations, the pitch angle of gimbal is limited within $[-90^{\circ},70^{\circ}]$ with camera FoV $[75^{\circ},55^{\circ}]$.
We set dynamic limits as ${\rm v}_{max}=2.0$ m/s, $\omega_{max}=1.0$ rad/s, ${\rm a}_{max}=1.0$ m/s$^2$, and ${\rm j}_{max}=0.5$ m/s$^3$ for all approaches.
The average comparison results are reported in Table.\ref{tab:benchmark}, where Exec. Time means trajectory execution time. 
We also visualize the comparisons in Fig.\ref{fig:qua}.

\begin{table}[H]
   \scriptsize
   \renewcommand\arraystretch{1.4}
   \tabcolsep=0.4mm
   \centering
   \vspace{-0.3cm}
   \caption{Ablation Study on The Proposed Planning Modules.  \label{tab:ablation}}
   \vspace{-0.2cm}
   \begin{tabular}{l|p{19pt}<{\centering}|p{19pt}<{\centering}|p{19pt}<{\centering}|p{18pt}<{\centering}|p{18pt}<{\centering}|p{18pt}<{\centering}|p{19pt}<{\centering}|p{19pt}<{\centering}|p{19pt}<{\centering}} 
   \hline
                       & \multicolumn{3}{c|}{\textit{Pipe}} & \multicolumn{3}{c|}{\textit{Christ the Redeemer}} & \multicolumn{3}{c}{\textit{Marina Bay Sands}} \\
   \cline{2-10}
   & \textbf{NR} & Ours & \textbf{GO} & \textbf{NR} & Ours & \textbf{GO} & \textbf{NR} & Ours & \textbf{GO} \\
   \hline
   \hline
  Comp. Time (ms) & \textbf{1461.5} & 1614.8 & 7126.7 & \textbf{397.2} & 506.7 & 899.5 & \textbf{843.2} & 954.3 & 4901.7 \\ 
  \hline
  Path Length (m) & 1480.1 & 1440.4 & \textbf{1381.9} & 581.2 & 530.4 & \textbf{493.4} & 1561.9 & 1396.2 & \textbf{1164.8} \\
   \hline
   \end{tabular}
   \vspace{-0.4cm}
\end{table}

\begin{table}[b]
   \scriptsize
   \renewcommand\arraystretch{1.4}
   \tabcolsep=0.3mm
   \centering
   \vspace{0.6cm}
   \caption{Computation Time of Each Module.  \label{tab:time}}
   \vspace{-0.2cm}
   \begin{tabular}{l|c|c|c|c|c} 
   \hline
                       & \multirow{2}*{SSD} & \multirow{2}*{Subspaces} & \multirow{2}*{\makecell[c]{Viewpoint \\ $\,$Generation$\,$}} & \multicolumn{2}{c}{Hierarchical Planning} \\
   \cline{5-6}
                       & & & & $\,$Safe Path Search$\,$ & Planning \\
   \hline
   \hline
   \textit{Pipe} & $\sim$80ms & 4 & $\sim$160ms & $\sim$1000ms & $\sim$200ms \\ 
   \hline
   \textit{Christ the Redeemer} & $\sim$35ms & 4 & $\sim$80ms & $\sim$210ms & $\sim$180ms \\
   \hline
   \textit{Marina Bay Sands} & $\sim$300ms & 10 & $\sim$150ms & $\sim$300ms & $\sim$200ms \\
   \hline
   \end{tabular}
   \vspace{-0.1cm}
\end{table}

From the evaluation, our method significantly outperforms others in computation efficiency, coverage completeness, and path quality.
Our framework achieves maximal coverage with minimal viewpoints due to the proposed skeleton-guided viewpoint generation.
Thanks to our effective SSD and parallel planning, our approach runs over 10x faster than existing methods.
For path quality, we generate a shorter coverage path and trajectory realizing less execution time.
Additionally, such advantages become more pronounced as the complexity and scale of the scene increase.

We perform ablation studies to demonstrate the effectiveness of local path refinement and hierarchical planning.
We denote our framework without local refinement as \textbf{NR}, while \textbf{GO} replaces hierarchical planning with global optimization of the path for all viewpoints together.
Table.\ref{tab:ablation} shows the contributions of local path refinement and hierarchical planning to path quality and computation efficiency.
Our method shortens the path with minimal additional computation compared to \textbf{NR}.
Fig.\ref{fig:refine} presents the effectiveness of local refinement.
Compared to \textbf{GO}, our approach markedly reduces computation time with minimal loss in path quality.
Moreover, we provide the runtime of each module and space decomposition details in Table.\ref{tab:time}.
Table.\ref{tab:time} also provides runtime of each module and space decomposition details.
Moreover, detailed performance information for each module can be found on our \href{https://hkust-aerial-robotics.github.io/FC-Planner}{project page}.

\begin{figure}[t]
	\begin{center}
      \vspace{0.2cm}
      \includegraphics[width=0.9\columnwidth]{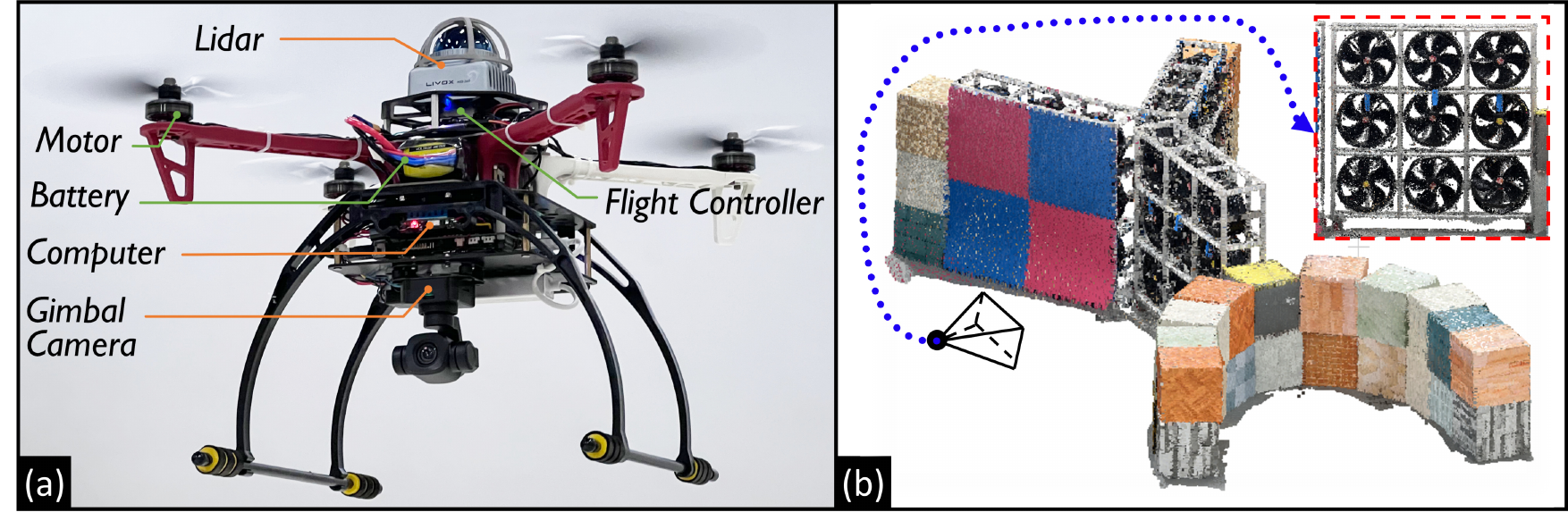}     
      \vspace{-0.6cm}
	\end{center}
   \caption{\label{fig:real-world} (a) The quadrotor with gimbal camera platform used for real-world coverage test. (b) The 3D reconstruction results after aerial coverage.
   }
   \vspace{0.1cm}
\end{figure}

\vspace{-0.3cm}
\subsection{Real-world Test}
\vspace{-0.1cm}

To further verify the proposed framework, we conduct a real-world coverage test in a challenging scene as shown in Fig.\ref{fig:top}(a), the size of which is \textbf{10}$\times$\textbf{5}$\times$\textbf{2.5} m$^3$.
For this test, we design a customized quadrotor as presented in Fig.\ref{fig:real-world}(a).
The gimbal rotates between $[-90^{\circ}, 20^{\circ}]$ and the camera FoV is $[60^{\circ}, 80^{\circ}]$.
The dynamic limits are set as ${\rm v}_{max}=0.7$ m/s, $\omega_{max}=1.0$ rad/s, ${\rm a}_{max}=0.5$ m/s$^2$, and ${\rm j}_{max}=0.5$ m/s$^3$.
We first collect the point cloud of the scene using Lidar.
Then, our framework produces the coverage trajectory as shown in Fig.\ref{fig:top}(b), where our method reasonably decomposes the scene into five subspaces for parallel planning.
Afterwards, the quadrotor captures images during executing the trajectory.
Lastly, we use these images to reconstruct the scene and the results shown in Fig.\ref{fig:real-world}(b) demonstrate our method indeed achieves complete coverage. 
More details about this test can be found in \href{https://www.youtube.com/watch?v=U-X4OddXI88}{our video}.

\vspace{-0.2cm}
\section{Conclusions}
\label{sec:conclusion}
\vspace{-0.1cm}

In this study, we present a skeleton-guided planning framework designed specifically for fast coverage of large and complex 3D scenes.
The proposed SSD efficiently extracts the scene's skeleton, enabling reasonable decomposition of the scene into simple subspaces and guiding efficient generation of specialized and informative viewpoints without pre-processing.
Based on SSD, the hierarchical coverage planner divides the entire planning problem into parallelizable sub-problems, combining the devised global planning and local refinement approaches to efficiently produce the high-quality coverage path.
The extensive experiments demonstrate the efficiency and superiority of our framework.
In the future, we plan to extend our framework to multi-UAV system for cooperative coverage in larger and unknown 3D scenes.

% \addtolength{\textheight}{-10.95cm}   % This command serves to balance the column lengths
%                                   % on the last page of the document manually. It shortens
%                                   % the textheight of the last page by a suitable amount.
%                                   % This command does not take effect until the next page
%                                   % so it should come on the page before the last. Make
%                                   % sure that you do not shorten the textheight too much.

\bibliography{fc}

\end{document}